\documentclass{article}

\usepackage[final,nonatbib]{nips_2018}

\usepackage[utf8]{inputenc} 
\usepackage[T1]{fontenc}    
\usepackage{hyperref}       
\usepackage{url}            
\usepackage{booktabs}       
\usepackage{amsfonts}       
\usepackage{nicefrac}       
\usepackage{microtype}      
\usepackage{amsmath}
\usepackage{euscript}
\usepackage{enumitem}
\usepackage{tikz}
\usepackage{tkz-tab}
\usepackage{caption}
\usepackage{latexsym}
\usepackage{amssymb}
\usepackage{amsmath}
\usepackage{subcaption} 

\title{Introducing Curvature to the Label Space}

\author{
  Conor ~Sheehan\thanks{These authors contributed equally to this work.} \\
  Dept. of Computer Science and Technology\\
  University of Cambridge\\
  15 JJ Thomson Avenue, Cambridge CB3 0FD.\\
  \And
  Ben ~Day\footnotemark[1]\\
  Dept. of Computer Science and Technology\\
  University of Cambridge\\
  15 JJ Thomson Avenue, Cambridge CB3 0FD.\\
  \texttt{benjamin.day@cst.cam.ac.uk} \\
  \AND
  Pietro ~Lio \\
  Dept. of Computer Science and Technology\\
  University of Cambridge\\
  15 JJ Thomson Avenue, Cambridge CB3 0FD.\\
}

\begin{document}

\maketitle

\begin{abstract}
One-hot encoding is a labelling system that embeds classes as standard basis vectors in a label space. Despite seeing near-universal use in supervised categorical classification tasks, the scheme is problematic in its geometric implication that, as all classes are equally distant, all classes are equally different. This is inconsistent with most, if not all, real-world tasks due to the prevalence of ancestral and convergent relationships generating a varying degree of morphological similarity across classes. We address this issue by introducing curvature to the label-space using a metric tensor as a self-regulating method that better represents these relationships as a bolt-on, learning-algorithm agnostic solution. We propose both general constraints and specific statistical parameterizations of the metric and identify a direction for future research using autoencoder-based parameterizations.
\end{abstract}

\section{Introduction}
Deep learning has had enormous success in pattern recognition tasks. Learning algorithms have surpassed human performance on image classification in some domains, for example. There is evidence that artificial neural networks, \textsc{ann}s, are able to accommodate learning hierarchical concepts \cite{Zeiler2013VisualizingNetworks} -- containing structures for identifying features shared by multiple categories of input, with specificity increasing into the network.

However, the prevailing method for labeling data -- one-hot encoding -- does not account  for hierarchical structure in the categories, nor any varying similarity between classes. This means that in the learning process, mistakes are treated as equally severe: confusing a dog with a wolf incurs the same `cost' as confusing a dog with a car. This suggests room for improvement in the learning process: if we could somehow communicate to the algorithm that it had the right idea when it confused a dog for a wolf, we might encourage the development of shared, transferable concepts, like `furry' or `four legs', within the network.

To address this problem we introduce curvature to the labeling scheme, a novel approach in the field. Whilst previous work has focused on the development of `hierarchical classifiers' \cite{Yan2014HD-CNN:Recognition,Zhu2017B-CNN:Classification,Roy2018Tree-CNN:Learning} that use additional handcrafted `coarse'-labels -- \textit{dogs} are \textit{mammals} are \textit{animals} -- and new network architectures to exploit the additional information, our approach uses no additional prior information and can be \textit{bolted-on} to any preexisting architecture that uses a one-hot encoding labeling scheme. Given the ubiquity of the scheme, this is a significant advantage of the approach. The mathematical basis that we build from has seen applications to deep learning in the past, treating the learning process as a path on a statistical manifold \cite{information-geometry}, but to the authors' knowledge there has been no work bringing curvature to the label space.

\begin{figure}
\centering
     \begin{subfigure}[b]{0.45\textwidth}
            \begin{center}
            \includegraphics[height=0.75\textwidth]{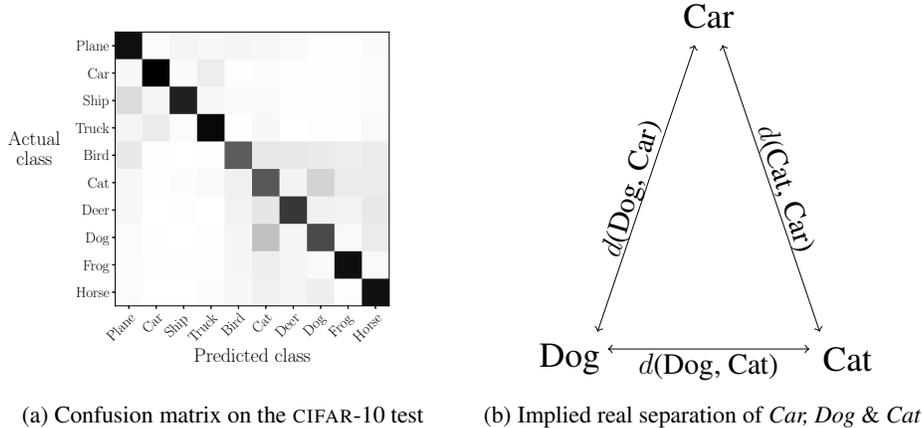}
            \end{center}
          \caption{Confusion matrix on the \textsc{cifar}-10 test}
          \label{fig:A}
     \end{subfigure}
     \begin{subfigure}[b]{0.45\textwidth}
          \centering
          \resizebox{0.75\linewidth}{!}{
            \begin{tikzpicture}
                \draw[<->] (0.9,0.1) node{} -- (2.3, 4.23) node{};
                \draw[<->] (1.05,-0.15) node{} -- (3.95, -0.15) node{};
                \draw[<->] (4.1,0.1) node{} -- (2.7, 4.23) node{};
                \node at (2.5, 4.65) {\Large Car};
                \node at (0.5, -0.3) {\Large Dog};
                \node at (4.5, -0.3) {\Large Cat};
                \node at (2.5, -0.4) {\large $d$(Dog, Cat)};
                \node at (1.4, 2.16) {\rotatebox{73.3}{\large $d$(Dog, Car)}};
                \node at (3.6, 2.16) {\rotatebox{288}{\large $d$(Cat, Car)}};
            \end{tikzpicture}}
          \caption{Implied real separation of \textit{Car, Dog \textup{\&} Cat}}
          \label{fig:B}
     \end{subfigure}
\caption{ \textbf{(a)} Heatmap of a confusion matrix on the \textsc{cifar}-10 dataset using the all-convolutional network \cite{Springenberg2014StrivingNet}. The mistakes made in prediction are those a human might make: in general vehicles are confused with other vehicles, top left block, and animals with other animals, bottom right block. Birds \& planes are the most frequently confused across the two main groups, likely a result of sharing a dominant background color (blue) and a similar cross-section. The most frequent mistake the network makes is classifying dogs as cats and vice-versa. \textbf{(b)} Schematic representation of distances in a curved label space that may result for the model that generated (a). $d$(Dog, Car) = $d$(Cat, Car) $>$ $d$(Dog, Cat). This arrangement improves upon the one-hot, equidistant encoding by recognizing the relative ease with which more closely related classes may be confused.}
\end{figure}

\section{Label space and loss functions}
In a one hot encoding scheme, the label for an example is a vector $\mathbf{y}$, with $y _\alpha$ equal to unity if the example belongs to class $\alpha$, and zero otherwise. The set of one hot vectors for all classes thus form an orthogonal, unit basis for a vector space we henceforth refer to as `label space'. We will use the word `class' to refer to all of the following: the class itself, the class index and the one hot vector corresponding to it. The output of a classifier $\mathbf{\hat{y}}$ is typically a probability vector in label space where $\hat{y}_\alpha$ is the estimated probability that the example belongs to class $\alpha$. We use the phrase `to confuse class $\alpha$ for class $\beta$' to mean estimating a high probability for class $\beta$ for an example belonging to class $\alpha$.

The loss functions typically used for classification tasks are functions of the Euclidean distance between the output and the label, $d_E(\mathbf{y}, \mathbf{\hat{y}}) = |\mathbf{\hat{y} - y}|$. For example mean squared error $\EuScript{L}_{\text{MSE}}$ and categorical cross entropy $\EuScript{L}_\text{CE}$ may be written as 
\begin{align}
\EuScript{L}_\text{MSE} &= \sum \limits _{\alpha}(\hat{y}_\alpha - y_\alpha)^2  = d_E ^2 (\mathbf{y}, \mathbf{\hat{y}}) \label{MSE_d}\\
\EuScript{L}_\text{CE} &= - \sum \limits _{\alpha} y_\alpha \log \hat{y}_\alpha = -\log \left( \frac{1}{2} \left[ 1 + |\mathbf{\hat{y}}|^2 - d_E^2(\mathbf{y}, \mathbf{\hat{y}}) \right] \right) \, . 
\end{align}
The squared distance between two one hot vectors $\boldsymbol{\alpha}$ and $\boldsymbol{\beta}$ is
\begin{align}
d^2_E(\boldsymbol{\alpha}, \boldsymbol{\beta}) = \begin{cases}
    \, 0 \quad & \boldsymbol{\alpha} = \boldsymbol{\beta} \\
    \, 2       & \text{else.}
\end{cases}
\end{align}
Confusing any two classes incurs the same loss, as the Euclidean distance between any two classes is the same.

\section{Curved label space}
We propose an alternative scheme in which the label space distances between classes are adjusted by the network automatically based on estimated cross-class similarity. We make use of a new distance function of the form
\begin{align}
d^2(\mathbf{\hat{y}}, \mathbf{y}) &= \sum \limits_{\alpha \beta} g_{\alpha \beta}  |\hat{y}_\alpha - y_\alpha| |\hat{y}_\beta - y_\beta|\, , \label{curved_distance}
\end{align}
where we have introduced a symmetric tensor $g_{\alpha\beta}$ which we will refer to as the metric tensor of the label space. In doing so we have created a curved label space; only in the case that the elements of $g_{\alpha\beta}$ are the elements of the identity matrix is the label space flat.

If we fix the diagonal elements of the metric tensor to unity, then the squared distance between two one hot vectors $\boldsymbol{\alpha}$ and $\boldsymbol{\beta}$ is
\begin{align}
d^2(\boldsymbol{\alpha}, \boldsymbol{\beta}) = \begin{cases}
    \, 0 \quad                & \boldsymbol{\alpha} = \boldsymbol{\beta} \\
    \, 2 (1+ g_{\alpha\beta}) & \text{else} \, .
\end{cases}
\end{align}

Thus the off-diagonal elements of the metric tensor allow the cross-class distances in label space to vary. We create `curved' loss functions by replacing Euclidean distances in common loss functions with the distance function in \eqref{curved_distance}, for example curved quadratic error (CQE) from MSE and curved crossentropy (CCE) from categorical crossentropy.
\begin{align}
\EuScript{L} _{\text{CQE}} &= \sum \limits_{\alpha \beta} g_{\alpha \beta}  |\hat{y}_\alpha - y_\beta| |\hat{y}_\beta - y_\beta| \\
\EuScript{L} _{\text{CCE}} &= - \log \left( \sum \limits _{\alpha\beta} g_{\alpha\beta} y_\alpha\hat{y}_\beta \right).
\end{align}
\section{Metric parameterization} \label{metpara}
Having outlined how to construct loss functions in a curved label space, we now turn to the task of parameterizing the metric in a way that reflects cross-class similarities. As mentioned earlier, the diagonals of the metric are set to unity:
\begin{align}
g_{\alpha\alpha} = 1  \quad \forall \alpha \, .
\end{align}
For the metric tensor to paramterize a meaningful distance representation, we require $g_{\alpha \beta} < g_{\alpha \gamma}$ if class \( \alpha \) is more similar to class \( \beta \) than it is to class \( \gamma \). A measure of distance, or effective distance, in label space may be derived from the confusion matrix, \( \mathbf{C} \), of an unbiased, well-performing classifier on the basis that more closely located centers will have greater overlap in the distribution of examples. Given that the existence of such a classifier would render the task solved, we substitute with the confusion matrix of the model being trained. Using the row normalized confusion matrix, \( \mathbf{P} \)\footnote{Denoted \( \mathbf{P} \) in reference to the diagonal elements of this matrix being the precision, also known as the positive predictive value}, protects against skew, whether in the dataset or model output, induced bias and provides a useful interpretation as
\begin{align}
P_{\alpha\beta} &= \dfrac{\text{number of class $\alpha$ classified as $\beta$}}{\text{number classified as $\beta$}} \\[1ex]
& \approx \text{likelihood an example classified as} \notag \\
&  \hspace{0.5cm} \text{a $\beta$ is an $\alpha$} \, . \notag
\end{align}
Using this measure, a matrix of effective class distances $S_{\alpha\beta}$ may be constructed as
\begin{align}
S_{\alpha\beta} = 1 - \tfrac{1}{2} (P_{\alpha\beta} + P_{\beta\alpha}) \, ,
\end{align}
as $S_{\alpha\beta}$ is smaller when classes $\alpha$ and $\beta$ are often confused, and is manifestly symmetric. We therefore set
\begin{align}
g_{\alpha\beta} = A S_{\alpha\beta} \quad \alpha \neq \beta.
\end{align}
where $A$ is a constant of proportionality, whose value relative to unity determines the importance of label space proximity relative to that of correct prediction.

We have yet to address exactly which confusion matrix is used to calculate $S_{\alpha\beta}$. We suggest an exponential moving average of previous batches' classifications during training, calculated recursively so as to reduce memory requirements as
\begin{align}
    \mathbf{\bar{P}}(t) &= (1-\alpha)\mathbf{\bar{P}}(t-1) + \alpha \mathbf{P}(t) \quad ; \quad \mathbf{\bar{P}}(0) = \alpha \mathbf{P}(0)
\end{align}
where $t$ is the epoch. This provides both smoothing to changes in the metric and a way of diminishing the relevance of errors made when the model parameters were significantly different or less well performing.

\section{Benchmarks}
Owing to space constraints, and the primary concern and contribution of this paper being to introduce the formalism of applying curvature to the label space, we discuss only our investigation of curvature for the benchmark \textsc{cifar}-10 and \textsc{cifar}-100 tasks \cite{Krizhevsky2009LearningImages}. We applied curvature to the label space of the All-Convolutional Network selected for its good performance, simple architecture and that the model had been applied to both tasks in the original paper \cite{Springenberg2014StrivingNet}. It is also our intention to avoid investigating curvature with novel architectures so as to avoid confounding the results of our investigations. No significant improvement in performance was found for this set up over using a flat label space. A simple transfer-learning task was also devised using these datasets where training was first carried out on the simpler \textsc{cifar-}10 task before being applied to the \textsc{cifar-}100, though again no significant improvement in performance was found when cross-fold validation was implemented.

\section*{Further Work}
Two branches of work are planned to further investigate the effects of curvature in the label space. The first will widen the scope of the work already completed using statistical parameterizations as described in section \ref{metpara} to include alternative models and tasks, particularly those believed to benefit from learning generalisable features such as transfer learning tasks. The second is to investigate parameterizing the curvature on the basis of an autoencoder's latent space representation. Separating the curvature from the model's own performance history has the obvious benefit of taking a reasonable form from the start of training and may also help to avoid biases that arise due to the network initialization. Relative separation in the latent space would also allow for a per-example based curvature and a curriculum learning-like training routine where examples believed to be more easily confused can be revisited and the loss controlled accordingly.


\bibliographystyle{unsrt}
\end{document}